\documentclass[11pt,letterpaper]{article}
\usepackage{emnlp2017}
\usepackage{times}
\usepackage{latexsym}
\usepackage{algorithm}
\usepackage[noend]{algpseudocode}
\usepackage{url}

\usepackage{color}
\usepackage{xcolor}
\definecolor{orange}{RGB}{255,127,0}
\newcommand{\cfbox}[2]{%
    \colorlet{currentcolor}{.}%
    {\color{#1}%
    \fbox{\color{currentcolor}#2}}%
}

\emnlpfinalcopy

\def\emnlppaperid{***}

\title{Boundary-based MWE segmentation with text partitioning}

\author{Jake Ryland Williams \\
  Drexel University \\
  30 N. 33rd Street \\
  Philadelphia, PA 19104 \\
  {\tt jw3477@drexel.edu} \\}

\date{}

\begin{document}

\maketitle

\begin{abstract}
  \fontsize{10}{11}\selectfont
  This work presents a fine-grained, text-chunking algorithm
  designed for the task of multiword expressions (MWEs) segmentation.
  As a lexical class, MWEs include a wide variety of idioms,
  whose automatic identification are a
  necessity for the handling of colloquial language.
  This algorithm's core novelty is its use of non-word tokens,
  i.e., boundaries, in a bottom-up strategy.
  Leveraging boundaries refines token-level information,
  forging high-level performance from relatively basic data.  
  The generality of this model's feature space
  allows for its application across languages and domains.
  Experiments spanning 19 different languages
  exhibit a broadly-applicable, state-of-the-art model.
  Evaluation against recent shared-task data 
  places text partitioning as the overall,
  best performing MWE segmentation algorithm,
  covering all MWE classes and multiple English domains
  (including user-generated text).  
  This performance,
  coupled with a non-combinatorial, fast-running design,
  produces an ideal combination for implementations at scale,
  which are facilitated through the release
  of open-source software.
\end{abstract}

\section{Introduction}
\label{sec:introduction}

{\bf Multiword expressions} (MWEs) constitute a mixed class
of complex lexical objects
that often behave in syntactically unruly ways.
A unifying property that ties this class together
is the lexicalization of multiple words into a single unit.
MWEs are generally difficult to understand through grammatical decomposition,
casting them as types of minimal semantic units.
There is variation in this non-compositionality property~\cite{bannard2003a},
which in part may be attributed to differences in MWE types.
These range from multiword named entities,
such as {\it Long~Beach,~California},
to proverbs,
such as {\it it~takes~one~to~know~one},
to idiomatic verbal expressions,
like {\it cut} it {\it out}
(which often contain flexible gaps).
For all of their strangeness they appear
across natural languages~\cite{jackendoff1997a,sag2002},
though generally not for common meanings,
and frequently with opaque etymologies
that confound non-native speakers.

\subsection{Motivation}

There are numerous applications in NLP for which
a preliminary identification of MWEs holds great promise.
This notably includes idiom-level machine translation \cite{carpuat2010a};
reduced polysemy in sense disambiguation \cite{finlayson2011a};
keyphrase-refined information retrieval \cite{newman2012a};
and the integration of idiomatic and formulaic language in learning environments~\cite{ellis2008a}.
Parallel to these linguistically-focused applications
is the possibility that MWE identification can positively affect
machine learning applications in text analysis.
Regardless of algorithm complexity,
a common preliminary step in this area is tokenization.
Having the ``correct'' segmentation of a text into words and MWEs
results in a meaning-appropriate tokenization
of minimal semantic units.
Partial steps in this direction have been taken
through recent work 
focusing on making the bag of phrases framework available
as a simple improvement to the bag of words.
However, that work~\cite{handler2016a} utilized only noun phrases,
leaving the connection between MWEs and
a comprehensive bag of phrases framework yet to be acknowledged.
With the specific focus of MWEs on idiomaticity,
a comprehensive bag of words and phrases framework would be possible,
provided the MWE identification task is resolved.

\subsection{Task description}
\label{sec:task}

Despite the variety that exist,
studies often only focus on a few MWEs classes,
or on only specific lengths~\cite{tsvetkov2011a}.
In fact, named entity extraction may be thought of as satisfying
the MWE identification task for just this one MWE class.
The problem has a broader framing when
all classes of MWEs are considered.
Furthermore, since a mixed tokenization of
words and phrases
as minimal semantic units
is a desired outcome,
it is helpful to consider this task
as a kind of fine-grained segmentation.
Thus, this work refers to its task as {\bf MWE segmentation},
and not identification or extraction.
In other words, the specific goal here is to
delimit texts into the smallest possible,
independent units of meaning.
\newcite{schneider2014a}
were the first to treat this problem as such,
when they created the first data set
comprehensively annotated for MWEs.
From this data set, an exemplar annotated record is:\footnote{
  Note that color/indices redundantly indicate separate MWEs,
  with the colored box highlighting an MWE's gap,
  and black, unnumbered text tokenized simply as words. 
}
\begin{center}
  My wife had \textcolor{red}{taken\textsubscript{1}}
  \cfbox{red}{her \textcolor{blue}{07'\textsubscript{2} Ford\textsubscript{2} Fusion\textsubscript{2}}}
  \textcolor{red}{in\textsubscript{1}} for a routine
  \textcolor{orange}{oil\textsubscript{3} change\textsubscript{3}}.
\end{center}
whose segmentation is an example of the present focus of this work.
Note that the present study focuses only on MWE tokens,
does not aim to approach the task of MWE class identification,
and does not attempt to disambiguate MWE meanings.
For detailed descriptions of these other
MWE-related tasks,
\newcite{baldwin2010a} provide an extensive discussion.

\subsection{Existing work}

The identification of MWEs and collocations is an
area of study that has seen notable focus in recent years
\cite{seretan2008a,pecina2010a,newman2012a,ramisch2015a,schneider2014b},
and has a strong history of attention
(both directly and through related work)
in the literature
\cite{becker1975a,church1990a,sag2002}.
It has become commonplace for approaches to
leverage well-studied machine learning algorithms
such as structured perceptrons~\cite{schneider2014b}
and conditonal random fields~\cite{constant2011a,hosseini2016a}.
The flexibility of these algorithms allow researchers to
mix a variety of feature types,
ranging from tokens to parts of speech to syntax trees.
Juxtaposed to these relatively-complex models
exist the simpler and more-heuristic~\cite{cordeiro2015a}.
Some rely singularly on MWE dictionaries,
while others incorporate multiple measures or are rule-based,
like those present in the suite available through mwetoolkit~\cite{ramisch2015a}
or jMWE~\cite{kulkarni2011a}.

MWEs have been the focus of considerable attention
for languages other than English, too.
Hungarian MWE corpora focusing on light verb constructions
have been under development for some time \cite{nagy2011a}.
In application to the French language,
part-of-speech tagging has seen benefit~\cite{constant2011a}
through awareness and relativity to MWEs.
Recently, \newcite{savary2017a}
conducted a shared task for the identification of verbal MWEs
with a data set spanning 18 languages
(excluding English).
While extending this area of work to a large variety of languages,
this task saw
notable multilingual algorithmic developments~\cite{saied2017a},
but did not approach
the identification of all MWE classes, comprehensively.
On the other hand, a SemEval 2016 shared task~\cite{schneider2016a}
covered English domains and all MWE classes,
bearing the greatest similarity to the present work.
In general, these shared tasks have all highlighted
a need for the improvement of algorithms.

\section{Algorithms}

\subsection{Text partitioning}
{\bf Text partitioning} is a physical model
developed recently~\cite{williams2015a} for fine-grained text segmentation.
It treats a text as a dichotomous squence,
alternating between word ($w_i$) and non-word ($b_i$) tokens:
$$(\cdots, b_{i-1}, w_i, b_i, w_{i+1}, b_{i+1} \cdots)$$
The key feature of text partitioning
is its treatment of non-word, i.e., ``boundary'', tokens.
Acting like glue, 
these may take one of
two distinct states, $s\in\{0,1\}$,
identifying if a non-word token is bound ($b_i^1$) or broken ($b_i^0$).
{\bf A non-word token in the bound state
binds words together.}
Thus, a text partitioning algorithm
is a function that
determines the states of non-word tokens.

In its original development,
text partitioning was studied simplistically,
with space as the only non-word token.
In that work,
a threshold probability, $q$, was set.
For each space, $b_i$, in a text,
a uniform random binding probability, $q_i$, would be drawn.
If $q_i > q$, $b_i$ would be bound,
and otherwise it would be broken.
As a parameter,
$q$ thus allowed for the tuning of a text
into its collection of words ($q=1$), clauses ($q=0$),
or, for any value, $q\in(0,1)$,
a randomly-determined collection of $N$-grams.
While non-deterministic,
this method was found to preserve word frequencies,
(unlike the sliding-window method),
and made possible the study of Zipf's law
for mixed distributions of words and $N$-grams.

The present work utilizes the parameter $q$
to develop a supervised machine learning algorithm
for MWE segmentation.
A threshold probability, $q$, is still set,
and the supervised component is
the determination of the binding probabilities
($q_i$) for a text's non-word tokens.
Provided a gold-standard, MWE-segmented text:
$$(\cdots, b^{s_{i-1}}_{i-1}, w_i, b^{s_i}_i, w_{i+1}, b^{s_{i+1}}_{i+1} \cdots)$$
let $f(w_i,b^{s_i}_i,w_{i+1})$
denote the frequency at which a boundary $b_i$
is observed between $w_i$ and $w_{i+1}$
in the state $s_i$.
Provided this, a binding probability is defined as:
$$q_i = \frac{f(w_i,b^{1}_i,w_{i+1})}{f(w_i,b^{1}_i,w_{i+1}) + f(w_i,b^{0}_i,w_{i+1})}.$$
This basic, $2$-gram text partitioning model
makes the binding probabilities
a function of boundaries
and their immediately-surrounding words.
In principle, this might be
extended to a more-nuanced model,
with binding probabilities
refined by larger-gram information.

\subsubsection{Extensions}

Some MWEs consist of non-contiguous spans of words.
These varieties are often referred to as ``gappy'' expressions,
an example of which is shown in Sec.~\ref{sec:task}.
Text partitioning may easily be extended to handle gappy MWEs
by instituting a unique boundary token, e.g., 
$$b = \_\_{\rm GAP}\_\_$$
that indicates the presence of a gap.
Since gappy MWEs are relatively sparse as compared to other MWEs,
a single gap-boundary token is used for all gap sizes.
This is designed for a flexible handling of variable gap sizes,
given the relatively small amount of gold-standard data
that is presently available.
However, this may in principle be refined
to particular gap-sized specifications,
possibly ideal for higher precision
in the presence of larger quantities of gold-standard data.

A number of MWE types,
such as named entities,
are entirely open classes.
Often occurring only once,
or as entirely emergent objects,
these pose a significant challenge
for MWE segmentation,
along with the general sparsity and size
of the current gold-standards.
For their inclusion in the gold-standard datasets
and the general quality of automated taggers,
part-of-speech (POS) information
may generally be leveraged to increase recall.
These data are utilized
in a parallel text partitioning algorithm,
swapping tokens for tags.\footnote{
  Note that this requires
  the inclusion of a special POS tag, e.g., ``SP'',
  for the space character.
}
Via two independent thresholds,
$q_{\rm tok}$ and $q_{\rm POS}$,
the combined algorithm merges candidate MWEs.

\begin{algorithm}
  \caption{
    Pseudocode for the longest first defined (LFD) algorithm.
    Here, a candidate MWE's tokens are pruned from left to right
    for the longest referenced in a training lexicon, $lex$.
    When no form is found in $lex$,
    the first token is automatically pruned,
    (accepting it as an expression),
    leaving the algorithm to start from the next.
    Note that the ``$\frown$'' symbol indicates a concatenation operation in line 10,
    where the current $form$ is placed onto the end of the $lexemes$ array.
  }
  \label{alg:LFD}
  \begin{algorithmic}[1]
    \Procedure{LFD}{$tokens$}
    \State $lexemes \gets (\cdot)$
    \State $N \gets {\rm length}(tokens)$
    \While{$N$}
    \State $indices \gets (N+1):1$
    \For {$i\:\textbf{in}\:indices$}
    \State $form \gets {\rm join}(tokens[0:i])$
    \State $remaining \gets tokens[i:N]$
    \If {$form\in\:lex\:\textbf{or not}\:i-1$}
    \State $lexemes \gets lexemes^\frown form$
    \If {${\rm length}(tokens) = 1$}
    \State $tokens \gets (\cdot)$
    \Else
    \State $tokens\gets remaining$
    \EndIf
    \State $\textbf{break}$
    \EndIf
    \EndFor
    \State $N \gets {\rm length}(tokens)$
    \EndWhile
    \State $\textbf{return}\:lexemes$
    \EndProcedure
  \end{algorithmic}
\end{algorithm}

\subsection{The longest first defined}
\label{sec:LFD}
In the presented form, text partitioning
only focuses on information immediately local
to boundaries (surrounding word pairs).
This has positive effects for recall,
but can result in lower precision,
since there is no guarantee that
a sequence of bound tokens is an MWE.
For example, if presented with the text:
\begin{center}
  ``I go for take out there, frequently.''
\end{center}
the segment {\it take out there} might be bound,
since {\it take out} and {\it out there}
are both known MWE forms,
potentially observed in training.
To balance this,
a directional, lookup-based algorithm is proposed.
Referred to as the {\bf longest first defined} (LFD) algorithm
(see Alg.~\ref{alg:LFD}),
this algorithm prunes candidates
by clipping off the longest known (MWE) references
along the reading direction of a language.
This requires knowledge of MWE lexica,
which may be derived from both gold-standard data
and external sources (see Sec.~\ref{sec:mat}).
Continuing with the example,
if the text partitioning algorithm
outputs the candidate, {\it take~out~there},
it would next be passed to the LFD.
The LFD would find {\it take~out~there} unreferenced,
and check the next-shortest (2-word) segments, from left to right.
The LFD would immediately find {\it take out} referenced, output it,
and continue on the remainder, {\it there}.
With only one term remaining,
the word {\it there} would then be trivially output
and the algorithm terminated.
While this algorithm will likely fail when confronted
with pathological expressions,
like those in ``garden path'' sentences,
e.g., ``The prime number few.'',
directionality is a powerful heuristic in many languages
that may be leveraged for increased precision.

\section{Materials}
\label{sec:mat}

\subsection{Gold standard data}
\label{sec:gold}

Treating MWE segmentation as a supervised machine learning task,
this work relies on several recently-constructed MWE-annotated data sets.
This includes the business reviews contained in the
Supersense-Tagged Repository of English with a Unified Semantics for Lexical Expressions,
annotated by Schneider et al. \shortcite{schneider2014a, schneider2015a}.
These data were harmonized and merged
with the Ritter and Lowlands data set
of supersense-annotated tweets~\cite{johannsen2014a}
for the SemEval 2016 shared task (\#10) on
Detecting Minimal Semantic Units and their Meanings (DIMSUM),
conducted by \newcite{schneider2016a}.
The DIMSUM data set additionally possesses
token lemmas and gold-standard part of speech (POS) tags
for the 17 universal POS categories.
In addition to the shared task training data
of business reviews and tweets,
the DIMSUM shared task resulted in the creation
of three domains of testing data,
which spanned business reviews,
tweets, and TED talk transcripts.
All DIMSUM data are comprehensive
in being annotated for all MWE classes.

To evaluate against a diversity of languages
this work also utilizes data produced by the multinational,
European Cooperation in Science and Technology's action group:
PARSing and Multiword Expressions within a European multilingual network
(PARSEME)~\cite{savary2015a}.
In 2017, the PARSEME group
conducted a shared task with data spanning 18 languages\footnote{
  While the shared task was originally
  planned to cover 21 languages,
  corpus release was only achieved for
  Bulgarian (BG),
  Czech (CS),
  German (DE),
  Greek (EL),
  Spanish (ES),
  Farsi (FA),
  French (FR),
  Hebrew (HE),
  Hungarian (HU),
  Italian (IT),
  Lithuanian (LT),
  Maltese (MT),
  Polish (PL),
  Brazilian Portuguese (PT),
  Romanian (RO),
  Slovene (SL),
  Swedish (SV),
  and Turkish (TR).
  No sufficiently available native annotators
  were found for
  English (EN),
  Yiddish (YI),
  and Croatian (HR).
  High-level data
  (including POS tags)
  were provided for all of the 18 languages,
  except BG, HE, and LT.
}~\cite{savary2017a},
focusing on several classes of verbal MWEs.
So, while the PARSEME data are not annotated for all MWEs classes,
they do provide an assessment against multiple languages.
However, the resources gathered for the 18 languages
exhibit a large degree of variation in overall size
and numbers of MWEs annotated,
leading to observable differences in identifiability.

The gold standard data sets were
produced with variations in annotation formats.
The DIMSUM data set utilizes
a variant of the beginning inside outside (BIO) scheme~\cite{ramshaw1995a}
used for named entity extraction.
Additionally, their annotations indicate which tokens are linked to which,
as opposed to the PARSEME data set,
which simply identifies tokens to indexed MWEs.
Note that this has implications to task evaluation:
the PARSEME evaluations can only assess
tokens' presence inside of specific MWEs,
while the DIMSUM evaluations
can focus on specific token-token attachments/separations.
Evaluations against the DIMSUM datasets
are therefore more informative of
segmentation, than identification.
Additionally, the DIMSUM data sets use lowercase BIO tags
to indicate the presence of tokens inside of the gaps of others.
However, the DIMSUM data sets
provide no information on the locations of spaces
in sentences, unlike the PARSEME data sets,
which do.
Since the present work relies on knowledge of spaces
to identify token-token boundaries for segmentation,
the DIMSUM data sets had to first be pre-processed
to infer the locations of spaces.
This is done in such a way as to preserve comparability
with the work others, (discussed in Sec.~\ref{sec:preprocessing}).

\subsection{Support data}

The gold-standard data sets
(DIMSUM, and PARSEME)
exhibit variations in size, domain, language,
and in the classes of annotated MWEs.
Ideally, each of these data sets would cover all MWE classes.
Since the English data sets do, and many are open classes
(e.g., the named entity class readily accepts new members),
gold standards cannot be expected to cover all MWE forms.
So, to produce segmentations that identify rare MWEs,
like those that occur once in the gold standard data,
this work relies on support data.
Note that because the PARSEME data set covers a restricted set of MWE types
(verbal MWEs, only),
it would likely not help to incorporate external sources.
Thus, the support data described below
are only used for the English language experiments,
i.e., the DIMSUM data sets.

Since this work approaches the problem as a segmentation task,
information is needed on MWE edge-boundaries.
Thus, support data must present MWEs in their written contexts,
and not just as entries in a lexicon.
Example usages of dictionary entries provide this detail,
and are leveraged from Wiktionary (data accessed 1/11/16) and Wordnet~\cite{miller1995a}. %% \newcite{wiktionary2016}
These exemplified dictionary entries help to fill gold standard data gaps,
but still lack many noun compounds and named entities.
Outside of dictionaries, MWEs such as these may be found in encyclopedias.
Thus, the Wikipedia hyperlinks
present in all Wikipedia (data accessed 5/1/16) articles are utilized. %% \newcite{wikipedia2016}
Specifically, the exact hyperlink targets are used (not the displayed text),
and without using any term extraction measures for filtering,
as opposed to the data produced by \newcite{hartmann2012a}.
This results in data that are noisy,
with many entities that may not actually be classifiable as MWEs.
However, their availability and broad coverage
offset these negative properties,
which is exhibited by this work's evaluation.

\section{Methods}

\subsection{Pre-processing}
\label{sec:preprocessing}

None of the gold standard data sets explicitly
identify the locations of spaces in their annotations.
This is a challenge for the present work,
since it focuses on word-word boundaries
(of which space is the most common)
to identify the separations between segments.
This turns out to not be an issue with the PARSEME data sets,
which indicate when a given token is not followed by a space.
However for the DIMSUM data sets,
the locations of spaces had to be inferred.
To resolve this issue, a set of heuristic rules are adopted
with a default assumption of space on both sides of a tokens.
Exceptions to this default include,
group openings (e.g., brackets and parentheses) and
odd-indexed quotes (double, single, etc.),
for which space is only assumed at left; and
punctuation tokens (e.g., commas and periods),
group closures (e.g., brackets and parentheses), and
even-indexed quotes (double, single, etc.),
for which space is only assumed at right.
While these heuristics will certainly not
correctly identify all instances of space,
they make the data sets more faithful to their original texts.
Furthermore, since the annotations and evaluation procedures
only focus on links between non-space tokens,
the data may be re-indexed during pre-processing
so as to allow for any resulting evaluation
to be comparable to those of the data set authors'
and shared task participants'.
Thus, the omission of space characters and their inference in this work
only negatively impacts text partitioning's evaluation.
In other words, if this work were applied to
annotated data that properly represents space,
higher performance might be exhibited.

\subsection{Evaluation}
\label{sec:evaluation}

It is reasonably straightforward to
measure precision, recall, and $F_1$
for exact matches of MWEs.
However, this strategy is unreasonably coarse,
failing to represent partial credit
when algorithms get only portions of MWEs correct.
Thus, the developers of the different gold standard data sets
have established other evaluation metrics
that are more flexible.
Utilizing these partial credit MWE evaluation metrics
provides refined detail into the performance of algorithms.
However, these are not the same across
the gold standard data sets.
So, to maintain comparability of the present results,
this work uses the specific strategies associated to each shared task.

In application to the PARSEME data sets,
precision, recall, and $F_1$
describe tokens' presence in MWEs.
Alternatively, DIMSUM-style metrics
measure link/boundary-based evaluations.
Specifically, this strategy checks if the links between tokens are correct.
Note that this latter (DIMSUM) evaluation is better aligned to
the formulation of text partitioning,
but leaves the number evaluation points at one fewer per MWE
than the PARSEME scheme.
Thus, PARSEME evaluations favor longer MWEs more heavily.

\begin{table*}[t!]
  \fontsize{9}{11}\selectfont
  \label{tab:results}
  \centering
  \begin{tabular}{|c|c|c|c|c|c|c|c|c|}
    \hline
    {\bf Experiment} & \bf LFD & \boldmath $q_{\rm tok}$ & \boldmath $q_{\rm POS}$ & \boldmath $P$ & \boldmath $R$ & \boldmath $F_1$ & \bf Rank & \boldmath $F_1$\bf-range \\
    \hline
    %% \multicolumn{9}{|c|}{STREUSLE}\\
    %% \hline    
    %% EN & N & 0.5 & 0.58 & 0.6128 & 0.539 & 0.5731 & 2/2 & 0.6253 \\
    %% EN & Y & 0.79 & 0.46 & 0.7637 & 0.5817 & 0.6603 & 1/2 & - \\
    %% \hline
    \multicolumn{9}{|c|}{\bf DIMSUM}\\
    \hline
    EN & N & 0.5 & 0.71 & 0.5396 & 0.5507 & 0.5451 & 3/5 & 0.1348 -- 0.5724\\
    EN & Y & 0.74 & 0.71 & 0.6538 & 0.5606 & 0.6036 & 1/5 & - \\    
    %% EN & Y & 0.74 & 0.57 & 0.6368 & 0.3883 & 0.4825 & 3/5 & - \\
    Tweets & N & 0.5 & 0.71 & 0.5897 & 0.5226 & 0.55542 & 3/5 & 0.1550 -- 0.6109\\
    Tweets & Y & 0.74 & 0.71 & 0.6667 & 0.5185 & 0.5833 & 3/5 & - \\
    %% Tweets & Y & 0.74 & 0.57 & 0.737 & 0.4095 & 0.5265 & 3/5 & - \\
    Reviews & N & 0.5 & 0.71 & 0.5721 & 0.5584 & 0.5626 & 1/5 & 0.0868 -- 0.5408\\
    Reviews & Y & 0.74 & 0.71 & 0.6742 & 0.5823 & 0.6249 & 1/5 & - \\
    %% Reviews & Y & 0.74 & 0.57 & 0.5775 & 0.3225 & 0.4139 & 3/5 & - \\
    TED & N & 0.5 & 0.71 & 0.3984 & 0.6108 & 0.4823 & 3/5 & 0.2011 -- 0.5714\\
    TED & Y & 0.74 & 0.71 & 0.5810 & 0.6228 & 0.6012 & 1/5 & - \\
    %% TED & Y & 0.74 & 0.57 & 0.5592 & 0.5090 & 0.5329 & 3/5 & - \\
    \hline
    \multicolumn{9}{|c|}{\bf PARSEME}\\
    \hline
    BG & N & 0.79 & N/A & 0.7071 & 0.5141 & 0.5954 & 2/3 & 0.5916 -- 0.6615 \\
    BG & Y & 0.83 & N/A & 0.8534 & 0.4309 & 0.5727 & 3/3 & - \\
    CS & N & 0.73 & 0.0 & 0.7849 & 0.6655 & 0.7203 & 3/5 & 0.2352 -- 73.65 \\
    CS & Y & 0.9 & 0.59 & 0.8363 & 0.6324 & 0.7202 & 3/5 & - \\
    DE & N & 0.82 & 0.0 & 0.5582 & 0.2788 & 0.3719 & 4/6 & 0.283 -- 0.4545 \\
    DE & Y & 0.98 & 0.78 & 0.6892 & 0.2010 & 0.3112 & 5/6 & - \\
    EL & N & 0.78 & 0.0 & 0.3931 & 0.3815 & 0.3872 & 5/6 & 0.3871 -- 0.4688 \\
    EL & Y & 0.99 & 0.66 & 0.5755 & 0.3314 & 0.4206 & 4/6 & - \\
    ES & N & 0.64 & 0.71 & 0.7473 & 0.4098 & 0.5293 & 2/6 & 0.3093 -- 0.5839 \\
    ES & Y & 0.99 & 0.71 & 0.7526 & 0.4371 & 0.5530 & 2/6 & - \\
    FA & N & 0.57 & 0.68 & 0.7040 & 0.8313 & 0.7624 & 3/3 & 0.8536 -- 0.9020 \\
    FA & Y & 0.93 & 0.68 & 0.7028 & 0.8266 & 0.7597 & 3/3 & - \\
    FR & N & 0.73 & 0.0 & 0.6589 & 0.3836 & 0.4849 & 4/7 & 0.1 -- 0.6152 \\
    FR & Y & 0.88 & 0.0 & 0.9045 & 0.3592 & 0.5142 & 3/7 & - \\
    HE & N & 0.78 & N/A & 0.5969 & 0.2107 & 0.3115 & 2/3 & 0.0 -- 0.313 \\
    HE & Y & 1.0 & N/A & 0.9714 & 0.1812 & 0.3056 & 2/3 & - \\
    HU & N & 0.97 & 0.66 & 0.7221 & 0.6612 & 0.6903 & 2/6 & 0.6226 -- 0.7081 \\
    HU & Y & 0.97 & 0.66 & 0.7208 & 0.6568 & 0.6873 & 3/6 & - \\
    IT & N & 0.85 & 0.0 & 0.5497 & 0.3174 & 0.4024 & 2/5 & 0.1824 -- 0.4357 \\
    IT & Y & 0.97 & 0.92 & 0.6503 & 0.2804 & 0.3919 & 2/5 & - \\
    LT & N & 0.79 & N/A & 0.6567 & 0.1803 & 0.2830 & 1/3 & 0.0 -- 0.2533 \\
    LT & Y & 1.0 & N/A & 0.6471 & 0.1352 & 0.2237 & 2/3 & - \\
    MT & N & 0.86 & 0.0 & 0.1591 & 0.1538 & 0.1564 & 2/5 & 0.0 -- 0.1629\\
    MT & Y & 0.98 & 0.0 & 0.2126 & 0.1138 & 0.1483 & 2/5 & - \\
    PL & N & 0.66 & 0.0 & 0.8962 & 0.5966 & 0.7164 & 2/5 & 0.0 -- 0.7274 \\
    PL & Y & 0.66 & 0.0 & 0.9623 & 0.5966 & 0.7366 & 1/5 & - \\
    PT & N & 0.79 & 0.0 & 0.7518 & 0.4921 & 0.5948 & 4/5 & 0.3079 -- 0.7094 \\
    PT & Y & 0.95 & 0.0 & 0.8717 & 0.4605 & 0.6027 & 3/5 & - \\
    RO & N & 0.71 & 0.0 & 0.8350 & 0.7850 & 0.8092 & 3/5 & 0.7799 -- 0.8358 \\
    RO & Y & 0.87 & 0.0 & 0.8766 & 0.7832 & 0.8272 & 2/5 & - \\
    SL & N & 0.7 & 0.0 & 0.6606 & 0.4504 & 0.5356 & 1/5 & 0.3320 -- 0.4655 \\
    SL & Y & 0.76 & 0.0 & 0.7192 & 0.3959 & 0.5107 & 1/5 & - \\
    SV & N & 1.0 & 0.95 & 0.0949 & 0.7771 & 0.1691 & 5/5 & 0.2669 -- 0.3149 \\
    SV & Y & 1.0 & 0.95 & 0.1013 & 0.7751 & 0.1792 & 5/5 & - \\
    TR & N & 0.87 & 0.0 & 0.3852 & 0.3706 & 0.3778 & 5/5 & 0.4550 -- 0.5528 \\
    TR & Y & 0.9 & 0.91 & 0.3814 & 0.4037 & 0.3922 & 5/5 & - \\
    \hline
  \end{tabular}
  \caption{
    \fontsize{9}{11}\selectfont
    Evaluation results,
    including data sets (Experiment);
    the LFD's application ({\bf LFD});
    token ({\boldmath $q_{\rm tok}$})
    and POS ({\boldmath $q_{\rm POS}$}) thresholds;
    precision ({\boldmath $P$}), recall ({\boldmath $R$}), and F-measure ({\boldmath \bf$F_1$});
    shared-task rank ({\bf Rank});
    and shared task $F_1$ ranges ({\boldmath \bf $F_1$-Range}).
    DIMSUM experiments spanned three domains:
    Twitter (Tweets), business reviews (Reviews), and TED talk transcripts (TED),
    with combined evaluation under EN.
    PARSEME language experiments are identified by
    ISO 639-1 two-letter codes.
  }  
\end{table*}

\subsection{Experimental design}

The basic text partitioning model
relies on the single threshold parameter, $q$,
and integration of POS tags relies on a second.
So, optimization ultimately
entails the determination of parameters
for both tokens, $q_{\rm tok}$, and and POS tags $q_{\rm POS}$.
To balance both precision and recall,
these parameters are determined
through optimization of the $F_1$ measure.
In the absence of the LFD,
$F_1$-optimal pairs, $(q_{\rm tok}, q_{\rm POS})$,
are first determined via a full parameter scan over
$$(q_{\rm tok}, q_{\rm POS})\in\{0, 0.01, \cdots, 0.99, 1\}^2.$$
For a given threshold pair,
LFD-enhancement can then only
increase precision, while decreasing recall.
So, subsequent optimization with the LFD
is accomplished through scanning values of 
$q_{\rm tok}$ and $q_{\rm POS}$
in the parameter space
no less than those previously determined
for basic, non-LFD model.

The different experiments were conducted
in accordance with the protocols established
by the designers of data sets and shared tasks,
and in all cases,
an eight-fold cross-validation was conducted for optimization.
Exact comparability was achieved
for the DIMSUM and PARSEME experiments
as a result of the precise configurations of
training and testing data from the shared tasks.
Moreover, since an evaluation script was provided for each,
metrics reported for DIMSUM and PARSEME experiments
are in complete accord with the results of the shared tasks.
For the DIMSUM experiments,
results should be compared to the open track
(external data was utilized),
and for the PARSEME experiments,
results should be compared to the closed track
(no external data was utilized).

\section{Results}

Evaluations spanning the variety
of languages (19, in total)
showed high levels of performance,
especially in application to English,
where there was a diversity of domains
(business reviews, Tweets, and TED talk transcripts),
along with comprehensive MWE annotations.
Moreover, these results were generally observed
for text partitioning both with, and without the LFD.
As expected, application of the LFD
generally led to increased precision.
While integration of POS tags
was found to generally improve MWE segmentation
in all English experiments,
this was frequently not the case
in applications to other languages.
However, this observation
should be taken with consideration
for the restriction to the fewer MWE classes
(verbal MWEs, only)
annotated in the PARSEME (non-English) shared task languages,
and additionally the fact that no external data were used.
Detailed results for all DIMSUM and PARSEME experiments
are recorded in Tab.~\ref{tab:results}.

For the DIMSUM experiments,
final parameterizations were determined as
$(q_{\rm tok}, q_{\rm POS}) = (0.5, 0.71)$
for text partitioning, alone,
and $(q_{\rm tok}, q_{\rm POS}) = (0.74, 0.71)$
for the LFD-enhanced model.
Comparing the base and LFD-enhanced models,
higher overall performance
was always achieved with the LFD
(increasing $F_1$ by as many as 12 points).
Including text partitioning in the
shared-task rankings (for a total of 5 models)
placed the LFD-enhanced model
first at all domains but Twitter,
for which third was reached
(though within 3 $F_1$-points of first).
However, combining all three domains
into a single experiment
placed the LFD-enhanced text partitioning algorithm as first,
making it the best-performing algorithm, overall.
In application to the user-reviews domain,
text partitioning maintained first-place status,
even without the LFD enhancement.
For all other domains
the base model ranked third.

For the PARSEME experiments,
final parameterizations varied widely.
This is not surprising,
considering the significant variation
in data set annotations and domains
across the 18 languages.
Additionally, POS tags were found
to be of less-consistent value
to the text partitioning algorithm,
particularly when the LFD was not applied.
Indeed, cross-validation of the base model
resulted in $q_{\rm POS} = 0$ as optimal
for 11 out of the 15 languages
where POS tags were made available.
However, cross-validation of the LFD-enhanced algorithm
resulted in only 6 parameterizations
having $q_{\rm POS} = 0$ as optimal.
First place status was achieved
for three out of the 18 languages
(LT, PL, and SL),
and for all languages
aside from SV and TR,
mid-to-high ranking $F_1$ values were achieved.\footnote{
  Note that anomalous MWEs were observed in the DE HU data sets,
  where large portions of the annotated MWEs consisted of only a single token.
  While the PARSEME annotation scheme includes multiword components
  that span a single token,
  e.g., ``don't'' in {\it don't talk the talk},
  those observed in DE and HU were found outside of the annotation format.
  This included $27.2\%$ of all MWEs annotated in the DE test records
  and $64.8\%$ of all in the HU test records.
  Since text partitioning identifies segment boundaries,
  it cannot handle these anomalous MWEs,
  unlike the models entered into the PARSEME shared task.
  So to accommodate these and maintain comparability, 
  a separate algorithm was employed.
  This simply placed lone MWE tags on tokens
  that were observed as anomalous
  $50\%$ or more of the time in training.
}
In contrast to the DIMSUM data sets,
application of the LFD improved $F_1$ scores
in only roughly half of the experiments.

\section{Discussion}

Evaluation against the comprehensively-annotated
English data sets has shown
text partitioning to be the current
highest overall ranking MWE segmentation algorithm.
This result is upheld for two out of the three
available test domains (business reviews and TED talk transcripts),
with a close third place achieved against data from Twitter.
This exhibits the algorithms general applicability across domains,
and especially in the context of noisy text.
Combined with the algorithm's fast-running
and non-combinatorial nature,
this makes text partitioning
ideal for large-scale applications
to the identification of colloquial language,
often found on social media.
For these purposes, the presented algorithms
have been made available as open-source tools
as the Python ``Partitioner'' module,
which may be accessed through Github\footnote{
  \href{https://github.com/jakerylandwilliams/partitioner}{https://github.com/jakerylandwilliams/partitioner}
}
and the Python Package Index\footnote{
  \href{https://pypi.python.org/pypi/partitioner}{https://pypi.python.org/pypi/partitioner}
} for general use.

Unfortunately, the PARSEME experiments did not
provide an evaluation against all types of MWEs.
However, they did exhibit the
general applicability of text partitioning across languages.
So, while the PARSEME data are not sufficient
for comprehensive MWE segmentation,
trained models have also been made available
for the 18 non-english languages
through the Python Partitioner module.
Across the 18 PARSEME shared-task languages
text partitioning's $F_1$ values were
found to rank as mid to high,
with the notable exception of SV.
However, the SV data is peculiar
in being quite small,
with its training set
smaller than its testing set.
However, models entered into the PARSEME shared task
achieved roughly twice the $F_1$ score for SV,
indicating the possibility that
text partitioning requires some critical mass of
training data in order to achieve high levels of performance.
Thus, for general increases in performance
and for extension to comprehensive MWE segmentations,
future directions of this work will likely
do well to seek the collection of
larger and more-comprehensive data sets.

As defined,
text partitioning is subtly different from a 2-gram model:
it focuses on non-word boundary tokens,
as opposed to just word-word pairs.
Because this algorithm
relies on knowledge of boundary token states,
it cannot be trained well on MWE lexica, alone.
Fort this model to achieve high precision,
boundaries commonly occurring as broken
must be observed as such,
even if they are necessary components of known MWEs.
Thus, the use of boundary-adjacent words for prediction
is a limitation of the present model.
This may possibly be overcome
through use of more distant words and boundaries.
However, since gold-standard data are still relatively small,
they will likely require significant expansion
before such models may be effectively implemented.
Thus, future directions with more nuanced text partitioning models
highlight the importance of generating more gold standard data, too.

%% \section*{Acknowledgments}
%% The author thanks Sharon Williams
%% for her ongoing support and thoughtful conversation,
%% Chaomei Chen
%% for his early inspiration of the project,
%% Andy Reagan
%% for his collaboration on the
%% Python Partitioner module,
%% and Giovanni Santia
%% for his thoughtful conversations.
%% The work was initially conducted
%% through a University of California, Berkeley
%% research position at the School of Information,
%% and was completed at Drexel University's
%% Department of Information Science
%% in the College of Computing and Informatics.

%% Do not number the acknowledgment section.

\bibliography{MWEs}
\bibliographystyle{emnlp_natbib}

\end{document}